\documentclass[twocolumn]{article}
\usepackage{xcolor}
\usepackage{amsmath, amssymb, amsfonts}
\usepackage{graphicx}
\usepackage[utf8]{inputenc}
\usepackage[T2A,T1]{fontenc} 
\usepackage{CJKutf8}
\usepackage{csquotes}
\usepackage{tipa}
\DeclareUnicodeCharacter{0264}{\textipa{[G]}} 
\usepackage[russian, greek, english]{babel} 
\usepackage{amsmath, amssymb} 
\usepackage{hanging}
\usepackage{booktabs}
\usepackage{algpseudocode}
\usepackage[backend=biber, style=authoryear]{biblatex} 
\usepackage{hyperref}
\usepackage{lipsum}
\usepackage{times}
\usepackage{multirow}
\usepackage{rotating}
\usepackage{geometry}
\usepackage{threeparttable} 
\usepackage{array} 
\usepackage{caption} 
\hypersetup{
    colorlinks=true,
    linkcolor=blue,
    citecolor=blue,
    urlcolor=blue
}

\title{\textbf{Missing the human touch? A computational stylometry analysis of GPT-4 translations of online Chinese literature}}
\author{
  Xiaofang Yao\textsuperscript{1}, Yong-Bin Kang\textsuperscript{2}, Anthony McCosker\textsuperscript{2} \\
  \textsuperscript{1}The University of Hong Kong \\
  \textsuperscript{2}Swinburne University of Technology \\
  \texttt{xiaofang.yao@hku.hk} 
}
\date{} 

\begin{document}

\maketitle

\begin{abstract}
Existing research suggests that machine translations of literary texts remain unsatisfactory. Such quality assessment often relies on automated metrics and subjective human ratings, with little attention to the stylistic features of machine translation. Empirical evidence is also scant on whether the advent of AI will transform the literary translation landscape, with implications for other critical domains for translation such as creative industries more broadly. This pioneering study investigates the stylistic features of AI translations, specifically examining GPT-4’s performance against human translations in a Chinese online literature task. Our computational stylometry analysis reveals that GPT-4 translations closely mirror human translations in lexical, syntactic and content features. As such, AI translations can in fact replicate the ‘human touch’ in literary translation style. The study provides critical insights into the implications of AI on literary translation in the posthuman, where the line between machine and human translations may become increasingly blurry.
\end{abstract}

\newenvironment{keywords}{\par\vspace{\baselineskip}\noindent\textbf{Keywords: }}{\par\vspace{\baselineskip}}


\section{Introduction}
The growing dominance of machine translation (MT) in the translation profession and beyond has become an irreversible trend. The sensational release of ChatGPT in 2023 and other Large Language Models (LLMs) further complicates the relationship between the machine and the human. In light of this, pioneering translation scholars remind us that we are now witnessing a ‘posthumanist turn’ in translation, which necessities a focus on the “increased interaction between translators and computers” \parencite[384]{obrien2020translation}. Translation faces a posthumanist reality, where “meaning emerges through our meaningful interaction with the technology and the world” \parencite[186]{carl2021enactivist} and “creativity arises out of the friction between the two (the human and the machine)” \parencite[299, \textit{brackets my addition}]{othomas2017humanum}. “It is no longer possible to postulate an anthropocentric, hierarchical order where the non-human world is passively subject to the interventions and manipulations of the human subject” \parencite[289]{cronin2020translation}. Posthumanist translation studies, as \textcite{lee2023artificial} argues, should attend to the assemblage of humans, machines, networks, texts, platforms, objects and spaces in the enactment of translation. An illustrative example is \textcite{fenoulhet2020relational}'s autoethnographic study of her own relationship to texts, computers, databases and dictionaries, which exemplifies the distributed agency and embodied subjectivity of the researcher’s identity as a human literary translator \parencite[cf.][]{gourlay2015posthuman}.

Another orientation of posthumanist translation studies focuses on the capabilities of MT and their alignment with human translators, which also represents the agenda of the current study. To re-evaluate the interaction between human and technology in translation practice, first of all, necessitates an understanding of the accuracy and fidelity of statistical language analysis \parencite{cope2024multimodal} and the expanding spectrum and distribution of skills and techniques associated with machine augmented translation \parencite{hayles1999how,lee2023artificial}. In this regard, it is widely acknowledged that MT’s lacking in accuracy and nuance is increasingly being offset by their capability to produce fast and cost-effective translations. Even in the literary genre, traditionally seen as the most challenging for MT, NMTs have registered significant quality improvement \parencite{toral2015machine,toral2018level}. It is thus no surprising that \textcite{othomas2017humanum} would envision a future where MT becomes sufficiently sophisticated and nuanced that “any work in any literature may potentially be accessible to anyone from any language” (292). Amidst these debates, evidence about the stylistic features and alignment of human and machine translated text is urgently needed.

Emerging studies show that LLMs like GPT-3 and GPT-4 exhibit impressive translation capabilities. For example, GPT-3 outperforms commercial-grade NMT systems by producing less literal and more fluent translations \parencite{thai2022exploring,raunak2023gpt}. ChatGPT competes well with NMTs for high-resource and natural spoken languages \parencite{hendy2023good}, and performs adequately for low-resource languages when pivot prompting is used \parencite{jiao2023chatgpt}. \textcite{aburayyash2024ai} finds that GPT-4 maintains the touch of humour and informal tone in translating from Egyptian comedy to American English. However, comparisons between LLMs and NMTs often rely on automated metrics, human evaluations, and error analysis, which emphasize sentence alignment. It is unclear how LLMs perform in areas where MT fails, such as creativity (novelty and acceptability) \parencite{guerberof2022creativity}, structural integrity \parencite{omar2020machine}, and the literary style \parencite{moorkens2018translators} and whether this performance aligns with the “human touch” of human translation.

In light of the evolving literary translation capabilities of the machine, this study investigates the stylistic features of GPT-4 translations compared to human translations of Chinese online literature. Understanding AI’s performance in literary translation is crucial for assessing its impact on literary translation in the posthuman era, with implications for other critical domains for translation including the creative industries more broadly. While existing stylistics studies typically focus on a small-scope analysis of a couple of classical literature works available in the public domain, this study leverages computer science and machine learning by applying a computational stylometry analysis to a big dataset of online Chinese novels and their human translations. In so doing, the study makes three important contributions to research on AI translation by 1) developing innovative prompts to enhance GPT-4’s context-based translation capability and address the sentence-based bias in MT, 2) building an open-access parallel corpus of Chinese online literature, human and GPT-4 translations, which responds to the inadequate attention to open-access research practice in literary translation studies, and 3) combining statistical and computational stylometry analysis to comprehensively assess the stylistic features of AI literary translations, which addresses the research gap in understanding the style of the machine translator.

\section{Style in AI literary translation}
Style of language is important to literature, as the aesthetics, rhetorics and poetics shape the literary text as much as its content. In literary translation, style is understood as “a kind of thumb-print that is expressed in a range of linguistic—as well as non-linguistic—features” \parencite{baker2000towards,kenny2020machine}. To capture the distinctive styles of literary translators, studies have used literary stylistics \parencite[e.g.][]{wu2024representing} and corpus stylistics \parencite[e.g.][]{li2011translation,ryu2023corpus}, focusing on linguistic features, such as personal pronouns, syntactic complexity, and connectives. Non-linguistic features are sometimes considered as well \parencite{cipriani2023literary}. For example, \textcite{huang2015style} comprehensively explores style in literary translation using a parallel corpus consisting of the English translations of modern and contemporary Chinese novels, employing both corpus statistics (e.g., type-token ratio, mean sentence length, frequencies of reporting verbs) and translator’s consistent, purposeful strategies for processing source texts as stylistic features. 

On the other hand, stylometry analysis quantifies the linguistic style of literary translations by examining stylistic elements that can be numerically measured, such as length of sentences, range of vocabulary, and the frequencies of certain words or word forms \parencite{gomez2018}. Computational stylometry analysis is widely used for authorship attribution based on quantitatively measured linguistic evidence \parencite{abbasi2008writeprints}. The underlying assumption is that individuals have idiosyncratic habits of language usage, resulting in stylistic similarities between texts written by the same person. This method therefore aims to determine the correct author of a new, previously unseen document by analysing specific stylometric features extracted from a set of texts with known authorship. By comparing the stylometric features of the new document to those of the known texts, this analysis can predict the likely authorship of the new document. 

While stylometry analysis is similar to corpus stylistics as both employ statistical methods for characterising linguistic features, the former is more often used for style comparisons. For example, \textcite{ping2024retranslated} compare the styles of two retranslations of the well-known classic Chinese novel, Journey to the West, using word lists, keyword analysis and function word analysis. The study distinguishes the two translators’ unique styles based on corpus statistics such as lexical density, sentence length and function word choices \parencite[cf.][]{fang2023seeing}. Similarly, \textcite{wu2022translated} explore stylistic connections between the translated Chinese Wuxia fiction and Western heroic literature based on calculations of stylistic indices such as average word length, dispersion of word length, moving average type-token ratio, verb–adjective ratio, average sentence length, among others \parencite[see also][]{haverals2022style}. Notably, in these studies, style is narrowly defined as the recurring textual features such as characters and words to allow quantitative computation.

Despite extensive research on human translators’ style, few studies explore the style of MT. One exception is \textcite{lee2022how}, which uses computational stylometry to differentiate machine (Google, Bing, and Papago) and human translations of modern Korean novels. The study proceeds from the posthuman assumption that machine translators have their authorship and distinct styles as much as human translators. Using Support Vector Machine as a classification model and n-grams (1, 2, 3) as stylistic features, the study suggests human and three machine translators display distinct styles. In a subsequent principal component analysis, C. Lee further identifies that machine translators tend to use generic content words and formulaic phrases. The constrained use of third-person pronouns and the biased use of masculine pronouns by machine translators leads to further questions regarding the reliability of machine literary translation. The study moves away from the quality-centric discussion on MT and showcases the potential of computational stylometry analysis in discriminating big datasets of human and machine literary translations. 

To date, research into how AI-generated (including LLM) literary translations differ stylistically from human translations remains sparse. This inquiry is crucial as emerging MT studies reveal that advanced AI models, such as the GPT series, often outperform NMTs by producing natural-sounding and fluent language outputs that more closely resemble human languages \parencite[e.g.][]{thai2022exploring,hendy2023good,jiao2023chatgpt,raunak2023gpt,aburayyash2024ai}. However, findings from related fields provide conflicting evidence. \textcite{zaitsu2023distinguishing}, for example, applies a stylometric analysis to distinguish ChatGPT-generated and human-written academic papers in Japanese. They focus on features such as parts-of-speech, postpositional particles, commas positioning, and the rate of function words. Their study found ChatGPT exhibits unique distributions in these features compared to human writers, suggesting a distinctive AI-generated textual style. Despite this promising evidence, it remains unclear how state-of-the-art AI compares to human translators in the domain of literary translation style.

A significant challenge in the area of literary translation is the limited access to comprehensive literary corpora for stylistic analysis. Existing research often relies on multilingual test datasets such as Flores-101 and WMT22 (19, 20, 21) \parencite[e.g.][]{thai2022exploring,jiao2023chatgpt}, or small, researcher-compiled, closed-access datasets of classic literature \parencite[e.g.][]{wu2022translated,wu2024representing}. This creates a sampling bias towards classic literature and highlights the need for more open-access approaches in the computational study of literary translation. 

In the context of Chinese-English literary translation, Chinese Internet literature, a thriving digital literary genre which attracts mass participation in literary writing, reading, criticism and adaptation \parencite{hockx2015internet}, has been largely overlooked. Often stereotyped as entertainment, Chinese web fiction parodies, extends and innovates the themes and languages of literature and addresses realistic issues such as gender identity, cultural ideology and popular discourse in digital cultures \parencite{feng2013romancing,shao2015wangluo,ouyang2023history}. Translations of these web novels are typically outsourced to agencies, crowdsourced from fan communities, or increasingly, automated with AI tools \parencite{tian2016fandom}. Despite the potential of AI in facilitating the interlingual and transcultural dissemination of popular fiction, scholarly research in this area remains limited \parencite[but see][]{cao2021accommodating,li2021transcultural,chang2022reader}.

Echoing \parencite{lee2022how}, the current study shifts its analytical focus from the quality of MT as often assessed by automated evaluation metrics (e.g., TER, BLEU, CHRF) and human evaluations \parencite{castilho2018approaches}, to examining the stylistic features of AI-generated literary translations. Using a statistical and computational stylometry approach, we investigate whether human and AI translations significantly differ in their stylistic features. Designating Chinese Internet literature as a literary translation task for the GPT-4 translator, this study addresses the following research questions: (1) Are there consistent stylistic similarities or differences between human and GPT-4 literary translations? (2) Which stylometric features, if any, can distinguish GPT-4 translations from human translations? 

\section{Methods and data} 
\subsection{Target large language model: GPT}
This study employed the GPT-4 (8k)\footnote{GPT-4 (8k) was the latest version available at the time of our analysis (Nov. 2023 – Feb. 2024).}, a state-of-the-art LLM, to translate Chinese novels into English. GPT-4 (8k) represented the latest iteration available in the Oceania region as of early 2024. This ensured that the most up-to-date LLM capabilities were tested. Our preliminary experiments showed that GPT-4 exhibited greater stability and reliability in translation outputs than its predecessor, GPT-3.5-turbo. This enhanced stability was crucial for maintaining the accuracy and consistency of AI-generated translations. However, GPT-4’s superior performance comes with higher usage costs than GPT-3.5. The need to balance between model quality and cost-effectiveness guides our decision on the size of the dataset for this study.

\subsection{Dataset: Online Chinese novels}
The dataset used for this study is a subset of the open-source BWB dataset\footnote{https://github.com/EleanorJiang/BlonDe?tab=readme-ov-file\#-the-bwb-dataset} \parencite{jiang2022blonde}. This large-scale document-level Chinese-English parallel dataset includes Chinese online novels across various genres along with their corresponding English translations sourced from the Internet. The translations are performed by translation agencies, addressing the limitation of previous studies that lacked extensive literary text datasets with human translations. The study by \textcite{jiang2023} focused on developing an automated evaluation metric by testing three NMT systems, leaving the distinct literary characteristics of machine translated Chinese novels unexplored. To fill this gap, we conducted a random quality sampling of 25 books from the five most popular sub-genres in the dataset: Fantasy, Romance, Hero, Sci-Fi, and Mystery. Table 1 provides a summary of the dataset used. We subsequently selected the first 100 chunks of 3 KB (3072) bytes from each book. The chunk size balances capturing sufficient textual context for meaningful analysis while ensuring computational efficiency and manageable processing times.

\subsection{Machine translation approaches}
For LLMs such as GPT-4, designing effective prompts is crucial for optimising translation outputs. Our approach to prompt engineering considered two key factors: domain and context. Drawing from insights from the literature \parencite{gao2023design,jiao2023chatgpt,peng2023towards,zhang2023prompting}, we included the genre of each book as a domain-specific detail. Further, contextual understanding is seen as a unique advantage of human translators \parencite{kruger2022translation}. To emulate this, we prompted GPT-4 to consider the purpose of translation and target audience, alongside the previous chunk and its English translation as context when translating the current chunk. As noted by \textcite{castilho2023online}, supplying preceding context in MT helps resolve lexical ambiguity, gender, and number issues. 

Accordingly, we developed and applied two prompt templates: sentence-to-sentence and context-based prompts. The former focuses on translating individual sentences from Chinese to English, while the latter considers the previous Chinese-English pair, aiming to maintain coherence and contextual accuracy across segments. Both prompts generated fluent translations, with the addition of domain information marginally improving accuracy in tenses and lexical choices \parencite[cf.][]{gao2023design}. Few-shot prompts \parencite{hendy2023good} were not considered in our study as they would complicate the context-based prompt design and further strain the token limit of using GPT-4. The two prompts are shown as follows: 

\begin{itemize}
    \item \textbf{Sentence-to-sentence prompt template: }\textit{Translate the text from Chinse to English by translating sentence-by-sentence. Purpose of translation: translate the text in literary style. Target audience: adolescents and young readers in their early 20s. Genre of the text: recently released online popular Chinese novel about '{genre}'.}
    \item \textbf{Context-based prompt template: }\textit{Translate the text from Chinse to English by considering the most previously translated text as context when translating the current text. Purpose of translation: translate the text in literary style. Target audience: adolescents and young readers in their early 20s. Genre of the text: recently released online popular Chinese novel about '{genre}'.}
\end{itemize}

These two prompts were used to generate GPT-4 translations for our compiled dataset of online Chinese novels. Table 2 shows an excerpt from the complete dataset, in which source text is lined up with three translations (HT: Human Translation, S2S: Sentence-to-Sentence Translation, CTX: Context-based Translation). Our initial observation suggests that the three translations align well with the source text. 

\subsection{Stylometry analysis}
Our stylometry analysis follows a two-step approach: identifying relevant stylometric features and then conducting a computational analysis to explore differences among translated texts based on these features. Drawing on established methods in the field \parencite{abbasi2008writeprints,lagutina2019survey,kumarage2023stylometric,kang2023,manabe2021}, we investigated three categories of stylistic features: lexical, syntactic, and content features (see Table 3). Lexical features are used to analyse the choice and frequency of words within the text. Syntactic features examine sentence structure and grammatical usage. Content features focus on the thematic and topical aspects of the text. By employing this comprehensive set of stylometric features, we aim to capture the stylistic nuances of both human and GPT-4 translations, thereby gaining deeper insights into the capabilities of GPT-4 in literary translation tasks.

We compared the stylometric outcomes among HT, S2S, and CTX translations using statistical significance and prediction quality. The statistical method was applied to lexical and syntactic features, which are quantifiable by frequency counts and ratios. For content features, we explored how the most frequent words (MFW) could predict the author of translations using Burrows’ Delta \textcite{burrows2002delta}. Building on the work of NN, we trained a Logistic Regression Model to classify unknown translations based on different types of common content words. We also modelled different numbers of common words, ranging from 100 to 1000, to gauge the impact of the number of common words on the model’s prediction accuracy.

\section{Results and analysis}
Our analysis suggests that the human translator employed a significantly broader range of vocabulary and crafted simpler sentence structures, compared to their GPT-4 counterpart. Nevertheless, both human and GPT-4 translations demonstrate similar stylistic preferences regarding word types. 

\subsection{Lexical features}
Table 4 presents the results from the one-way ANOVA test for lexical features. It is evident that human translations are not only richer in character and word counts but also exhibit a more diverse vocabulary, which renders human translations more elaborate than those produced by GPT-4. This observation echoes existing research on translator strategies, which documents that professional translators routinely use explication to enhance text clarity \parencite{sorbo2018jane}. Notably, HT outperforms CTX but is on par with S2S in terms of vocabulary range. This finding suggests that while prompting GPT-4 to incorporate the context may slightly increase translation length, it tends to make vocabulary choices more uniform compared to S2S.

Across the 25 books, the HT of Book 22 demonstrates considerably higher vocabulary richness (see Table 5). Expressions such as “peer through ... the azure blue sky” in HT are notably more varied than the simpler “see the blue sky” in GPT-4 translations. Additional distinctions include “imitated his posture” (HT) versus “imitating him” (S2S) or “following his example” (CTX); and “was woven of rattan” (HT) versus “was made of bamboo” (S2S, CTX). GPT-4 appears to favour more common word choices, corroborating C. Lee’s (2022) finding that the machine translator tends to overuse basic standard language, which can result in a ‘bland’ style (Moorkens et al. 2018). There are also instances of similar vocabulary between human and AI translations, such as “a breeze blew into” (HT) and “floral fragrance” (CTX).

While equivalence is not the primary focus in stylometric analysis, this example reveals critical semantic errors in GPT-4 translations. In the sentence enclosed by square brackets, only HT accurately conveys that it is the roof, not the blue sky, that blocks the sun. HT also successfully resolves the ambiguity of the “arched goldfish-patterned roof”, whereas GPT-4 erroneously refers to an implausible “goldfish house”. This highlights a significant limitation of AI translation in ensuring semantic equivalence, especially in cases of ambiguous meanings.

\subsection{Syntactic features}
At the syntactic level, human and AI translations become more difficult to distinguish. Statistical analysis ascertains that features such as sentence length, function words, transition words and punctuation per character do not effectively differentiate between the two (see Table 6). The primary syntactic distinction lies in sentence types. Human translators tend to prefer simpler sentences compared to the AI, though this difference is only statistically significant between HT and CTX, and not between HT and S2S. In essence, S2S is comparable to HT across all sentence-level features. 

The translations of Book 296, which include a high proportion of complex sentences, provide insight into the syntactic style disparities between human and machine translators (see Table 7). Consider the sentences highlighted in blue. HT clearly opted for simpler sentence and conjunction structures: “… he slanted his body to look down. The cliff was steep and filled with protruding rocks”. In contrast, GPT-4 translations use more transposition strategies that involve changes in parts of speech from ST as well as the use of subordinate clauses: “Looking down, he saw a steep precipice, protruding rocks, and a deep, unfathomable valley” (S2S, participle clause) and “Looking down, he saw the cliff was steep, with protruding rocks and a deep valley that was bottomless” (CTX, prepositional clause). The frequent use of participle clauses appears to be a defining feature of S2S (see texts in red). With 4 out of 5 sentences constructed using participle clauses, GPT-4 translations are arguably more economical than HT. 

\subsection{Content features}
Content features were unable to reliably predict the authorship of translations. In other words, GPT-4 translations are not easily distinguishable by their content style, and this finding remains robust across our modelling of MFWs from 100 to 1000 in number. Figure 1 summarises the best predication accuracies across all content features. The model can only correctly attribute authorship to unknown translations with approximately 60\% accuracy when using nouns as a predictor and 55\% accuracy when using adjectives (see Figures 2 and 3 respectively). The probability of accurately predicting machine translations based on other content-level stylometric features is nearly equal to chance (50\%). Essentially, our modelling analysis shows that human and GPT-4 translations cannot be confidently discriminated based on content features.

\begin{figure}
    \centering
    \includegraphics[width=1\linewidth]{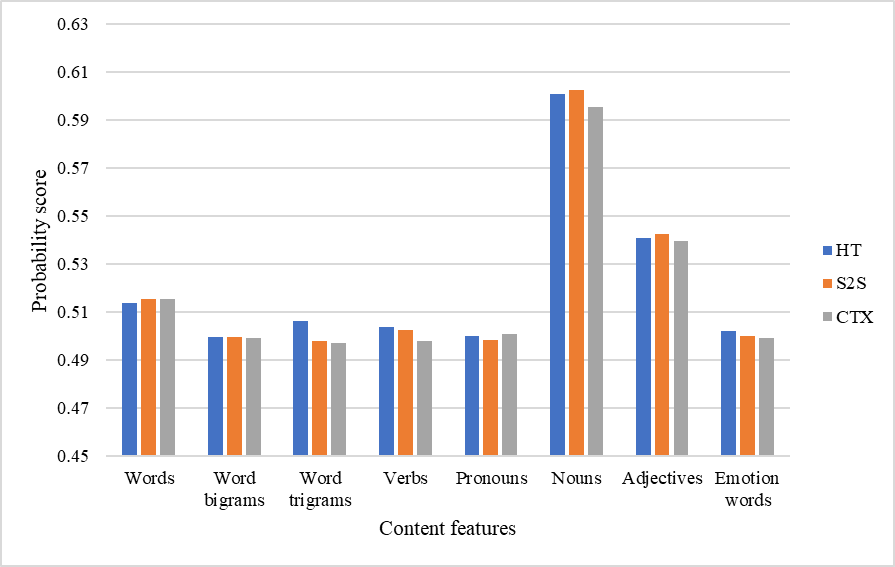}
    \caption{Prediction accuracy on content features}
\end{figure}

\begin{figure}
    \centering
    \includegraphics[width=1\linewidth]{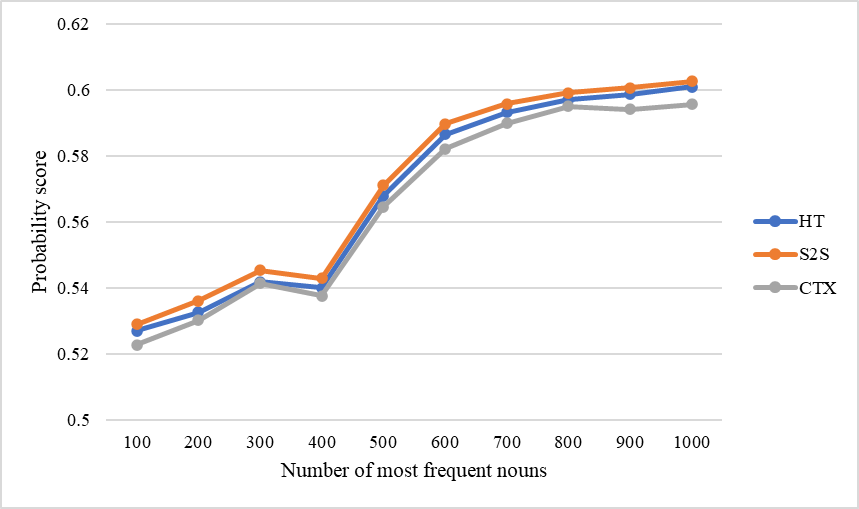}
    \caption{Prediction accuracy, modelling from 100 to 1000 nouns }
\end{figure}

\begin{figure}
    \centering
    \includegraphics[width=1\linewidth]{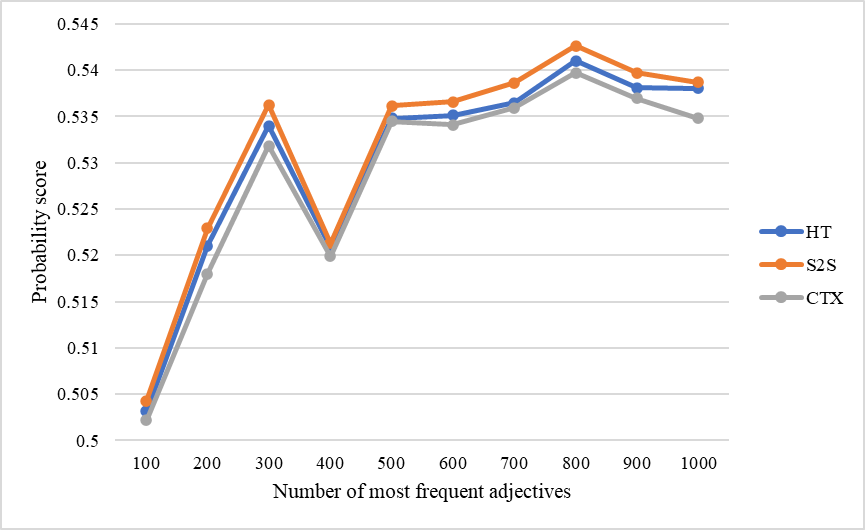}
    \caption{Prediction accuracy, modelling from 100 to 1000 adjectives }
\end{figure}

Nevertheless, the spike in probability scores for nouns and adjectives merits further attention. Table 8 provides an extended excerpt from the dataset, revealing a few observable differences in noun choices, for instance, “Spirit Firmament Door” (HT) versus “Lingxiao Palace” (S2S and CTX). In the translation of proper names, the human translator opted for translation, whereas the machine translator used a transliteration technique. This tendency to transliterate may indicate the machine translator’s incompetence in disambiguating the semantic meanings of culturally rich concepts in the ST. In HT, “for thou to use” infuses the text with an ancient flair whereas such dynamic equivalence is not achieved by GPT-4 translations. In addition, a noticeable error by the GPT-4 translator involves the homograph “\begin{CJK*}{UTF8}{gbsn}纥\end{CJK*}” (inferior silk), which should be translated as /x\textipa{G}35/ in person’s names and /k\textipa{G}55/ when referring to a knot with thread. This underscores the challenges that LLMs such as GPT-4 face in processing Chinese heteronyms. 

\section{Discussion and conclusion}
This study contributes to emerging research exploring AI translation styles, assessing the stylometric features of human and GPT-4 translations in the context of Chinese Internet literature. Our statistical and modelling analysis suggests that human translations stand out in the number and variety of words they use and exhibit simpler sentence structures. Meanwhile, their stylistic choices converge with those of the AI translator across all content features and most lexical and syntactic measures. Given the similarities between human and GPT-4 translations across a comprehensive set of stylistic metrics in our case study of translations of Chinese Internet literature, we may need to reconsider the beliefs that machine translations are ‘missing the human touch’ stylistically and that human translations represent the gold standard in the literary genre. While further research across different contexts is needed, our findings have implications for assessing the trustworthiness of GPT models in translation practice. 

Despite these promises, our analysis also reveals several inadequacies in GPT-4 translations, such as difficulty in comprehension of ambiguous texts, repetitive sentence structures, over-reliance on transliteration of proper nouns, and incorrect rendition of heteronym and/or polysemy. It appears that the GPT-4 translator still struggles to achieve equivalence between the source and target texts due to its inability to fully comprehend the semantic meanings of ambiguous Chinese texts. Although GPT-4 translations are comparable to human translations on syntactic stylometry measures, they exhibit eccentricities in sentence structure, particularly an overuse of dependent clauses. The translation of proper names involves an optional translation shift since there is no lexical and cultural equivalent in English, and the GPT-4 translator often opts for transliteration as opposed to the literal translation adopted by the human translator. The translation of Chinese polysemy and heteronym presents significant challenges for GPT-4. 

In addition to the empirical findings, this study has made two significant contributions to posthumanist translation studies. Firstly, through prompt engineering, it developed a set of reusable prompts configured with information about translation purpose, audience, genre and format, which can effectively moderate the translation output of GPT-4. Notably, this study proposed an innovative approach to eliciting context-based translation by including the previous source-target pair as input before initiating the translation of the next segment of source text. The prompt engineering process thus bridges insights from both functionalist translation studies and MT studies. Secondly, built upon the BWB online Chinese novel corpus, this study curated a parallel corpus of Chinese novels, their human translations, and GPT-4 translations, matched at the segment level. This corpus will be open access for reuse in further research, such as dependency parsing of human and machine translations to reveal structural differences, or for translator training.

The fidelity and overall plausibility of the GPT-4 translation are remarkable as it closely mirrors human translation across all the stylometric measures. However, the subtle differences and challenges highlight the persistent important collaborative role of specialist translators, especially in culturally sensitive contexts such as literature and the creative industries. \textcite{obrien2024human} note that machines should not be seen as threats to the human activity of translation, referring to this as the “dualism of humans and machines” (392). With the increasing reliance on MT in both professional and educational settings \parencite[e.g.][]{ji2023translation}, particularly for large-scale content translation such as online literature, our findings provide insights into where LLMs can support human translators. The subtle stylistic differences identified in our study, however, may prove more critical in contexts such as health or legal text translation. “While NMT represents a significant shift in computational terms, it has not changed the role or position of the translator in the translation value chain” \parencite[151]{ragni2022what}. Handling ambiguity or proper nouns in those translation contexts can be more consequential, and hence require specialist human oversight.

This study has practical significance for both the translation industry and the academic community by laying the groundwork for potential applications of AI translation models like GPT-4. Following on \textcite{hayles1999how}'s theoretical take on posthumanism, human translation can be viewed as part of a distributed system in which human capability is inherently embedded in non-human networks including AI tools such as GPT-4. We thus align with posthumanist scholars and advocate for rethinking the human factor in translation technology development, where machines may augment the volume, accessibility, and range of human translations. The reusability of prompts and the context-based translation approach presented in this study provide a mechanism for refining machine translation outputs, making it more adaptable to specific contexts and audiences. Our study also offers a way to consistently measure stylistic features of machine outputs. These contributions may have implications for improving translation workflows, reducing human intervention in routine tasks, and optimising machine-assisted translation. Amid the enthusiasm surrounding GPT-4 as the state-of-the-art in translation capabilities and its close alignment with human language, future studies should continue this research momentum and further interrogate the interaction between human and machine capabilities from both computational and sociological perspectives. 

\printbibliography 

\section*{Acknowledgments}
This study is part of the "Critical Capabilities for Inclusive AI" project funded by the ARC Centre of Excellence for Automated Decision Making and Society (CE200100005).

\onecolumn
\appendix
\section{Appendix}

\begin{table}[ht]
\centering
\caption{The 25 Chinese novels selected from the BWB dataset for analysis in the study}
\begin{tabular}{>{\raggedright\arraybackslash}p{0.1\linewidth}>{\raggedright\arraybackslash}p{0.5\linewidth}>{\raggedright\arraybackslash}p{0.25\linewidth}}
\hline
\textbf{Book ID in BWB} & \textbf{Book title} & \textbf{Genre} \\
\hline
3 & The Divine Nine-Dragon Cauldron & Mysterious Fantasy \\
22 & Way of the Devil & Mysterious Fantasy \\
192 & Destroyer of Ice and Fire & Mysterious Fantasy \\
116 & The Path Toward Heaven & Mysterious Fantasy \\
229 & Legend of Ling Tian & Mysterious Fantasy \\
127 & Perfect World & Fantasy \\
296 & Rise of Humanity & Fantasy \\
328 & Shen Yin Wang Zuo & Fantasy \\
338 & The Magus Era & Fantasy \\
287 & Hedonist Sovereign & Fantasy \\
14 & Strongest Abandoned Son & Hero \\
24 & My Master Disconnected Yet Again & Hero \\
15 & Reincarnation Of the Strongest Sword God & Hero \\
128 & Renegade Immortal & Hero \\
157 & Gate of God & Hero \\
63 & Hello, Mr. Major General & Romance \\
275 & My Beautiful Teacher & Romance \\
121 & Endless Pampering Only for You & Romance \\
168 & The King of Hell's Genius Pampered Wife & Romance \\
314 & The Strongest System & Romance \\
5 & Monster Paradise & Science Fiction \\
177 & Miracle Throne & Science Fiction \\
323 & Tales of the Reincarnated Lord & Science Fiction \\
292 & My MCV and Doomsday & Science Fiction \\
380 & Warlock of the Magus World & Science Fiction \\
\hline
\end{tabular}
\end{table}

\begin{table}[ht]
\centering
\caption{Comparison of Translations for Sample Text}
\begin{tabular}{>{\raggedright\arraybackslash}p{0.2\linewidth}>{\raggedright\arraybackslash}p{0.7\linewidth}}
\hline
\textbf{Translations} & \textbf{Sample text from dataset (3.chs re.txt)} \\
\hline
ST & \begin{CJK*}{UTF8}{gbsn}秦国公见苏羽陷入回忆，只当他难以割舍对姜雪晴情谊，叹息道："枫儿年轻不懂事，他的过错，我愿意付出赔偿，你想要什么，尽管提，但，作为父亲，我希望你能成全枫儿，他对雪晴姑娘，是认真的。\end{CJK*}" \\
HT & As Duke Qin watched Su Yu recall his past, he saw that Su Yu could not give up his friendship with Jiang Xueqing. He sighed, “Feng-er is still young and naïve. I am willing to compensate you for his mistakes. You can ask for whatever you want. However, as his father, I hope you can give your blessings to Feng-er. He is really serious with Jiang Xueqing.”  \\
S2S & Seeing Su Yu lost in memories, Duke Qin assumed he was struggling to let go of his feelings for Jiang Xueqing. He sighed, ‘Feng is young and naive. I am willing to compensate for his mistakes. You can ask for anything you want. But as a father, I hope you can let Feng be. He is serious about Miss Xueqing.’  \\
CTX & Seeing Su Yu lost in memories, Duke Qin assumed he was struggling to let go of his feelings for Jiang Xueqing and sighed, ‘Feng’er is young and ignorant. I’m willing to compensate for his mistakes. You can ask for anything you want. But as a father, I hope you can let Feng’er have his way. He is serious about Miss Xueqing.’ \\
\hline
\end{tabular}
\end{table}

\begin{table}[ht]
\centering
\caption{Range and definition of stylometric features}
\begin{tabular}{>{\raggedright\arraybackslash}p{0.15\linewidth}>{\raggedright\arraybackslash}p{0.2\linewidth}>{\raggedright\arraybackslash}p{0.6\linewidth}}
\toprule
\textbf{Type of features} & \textbf{Specific metrics} & \textbf{Definitions} \\
\midrule
\multirow{7}{*}{Lexical} & Character count & Total length of text strings \\
 & Total words & Total number of words \\
 & Character per word & Average number of characters per word \\
 & Vocabulary richness & The number of unique words divided by the total number of words \\
 & Word length & Frequency distribution of word lengths (ranging from 1 to 15 characters) \\
 & Common bigrams & The ratio of each common bigram, such as \textit{'th', 'he', 'in', 'er', 'an', }etc, to the total number of the bigram occurrences \\
 & Common trigrams & The ratio of each common trigram, including \textit{'the', 'and', 'ing', 'her', 'hat', }etc, to the total number of the trigram occurrences \\
\multirow{5}{*}{Syntactic} & Sentence length & The total number of words divided by the total number of sentences \\
 & Punctuation per character & Average frequencies of punctuations over the total character count \\
 & Function words & Average frequencies of function words (such as \textit{'the', 'and', 'a', 'to', 'of', }etc.) over the total number of words \\
 & Transition words & Average frequencies of transition words (such as \textit{'after', 'however', 'because', 'although', 'first', }etc.) over the total number of words \\
 & Simple/complex sentence types & The ratio of simple and complex sentences, where a \textit{simple sentence} contains an independent clause, and a \textit{complex sentence} contains one independent clause and at least one dependent clause \\
\multirow{8}{*}{Content} & Words  & Individual words \\
 & Word bigrams & Two words, such as \textit{'his\_eyes', 'nalan\_mansion', 'her\_face', 'he\_was', 'the\_third', }etc. \\
 & Word trigrams & Three words, such as \textit{'the\_underworld\_king', 'of\_spiritual\_power', 'at\_this\_time', 'the\_spiritual\_energy', 'men\_in\_black', }etc. \\
 & Pronouns & \textit{'her', 'she', 'you', }etc. \\
 & Nouns & \textit{'city', 'people', 'spirit', }etc. \\
 & Verbs & \textit{'shattered', 'become', 'holding', }etc. \\
 & Adjectives & \textit{'ordinary', 'large', 'few', }etc. \\
 & Emotion words & Words that are labelled as \textit{'positive', 'negative', 'fear', 'anticipation', 'anger', }etc. \\ \bottomrule

\end{tabular}
\end{table}

\begin{sidewaystable}
\begin{threeparttable}
\caption{ANOVA results for lexical features}
\begin{tabular}{ll l l l l l l l l l l l}
\hline
\multirow{2}{*}{\textbf{Lexical features}} & \multicolumn{2}{c}{\textbf{HT}} & \multicolumn{2}{c}{\textbf{S2S}} & \multicolumn{2}{c}{\textbf{CTX}} & \multirow{2}{*}{\textbf{F}} & \multirow{2}{*}{\textbf{Sig.}} & \multicolumn{4}{l}{\textbf{Pair-wise comparison}} \\
 & M & Std. & M & Std. & M & Std. &  &  & Group 1 & Group 2 & Mean dif. & Sig. \\
\hline
\multirow{3}{*}{Total character counts} & \multirow{3}{*}{406751.200} & \multirow{3}{*}{33851.595} & \multirow{3}{*}{357258.400} & \multirow{3}{*}{22377.167} & \multirow{3}{*}{363553.800} & \multirow{3}{*}{22477.917} & \multirow{3}{*}{25.298} & \multirow{3}{*}{0.000*} & CTX & HT & 43197.400 & 0.000* \\
 &  &  &  &  &  &  &  &  & CTX & S2S & -6295.400 & 0.685 \\
 &  &  &  &  &  &  &  &  & HT & S2S & -49492.800 & 0.000* \\
\multirow{3}{*}{Total words} & \multirow{3}{*}{72681.880} & \multirow{3}{*}{6047.358} & \multirow{3}{*}{64644.080} & \multirow{3}{*}{3861.915} & \multirow{3}{*}{66161.400} & \multirow{3}{*}{4007.840} & \multirow{3}{*}{20.250} & \multirow{3}{*}{0.000*} & CTX & HT & 6520.480 & 0.000* \\
 &  &  &  &  &  &  &  &  & CTX & S2S & -1517.320 & 0.499 \\
 &  &  &  &  &  &  &  &  & HT & S2S & -8037.800 & 0.000* \\
Average character per word & 4.597 & 0.142 & 4.604 & 0.096 & 4.564 & 0.101 & 0.857 & 0.429 & \multicolumn{4}{l}{--} \\
\multirow{3}{*}{Vocabulary richness} & \multirow{3}{*}{0.080} & \multirow{3}{*}{0.009} & \multirow{3}{*}{0.078} & \multirow{3}{*}{0.008} & \multirow{3}{*}{0.072} & \multirow{3}{*}{0.007} & \multirow{3}{*}{6.603} & \multirow{3}{*}{0.002*} & CTX & HT & 0.008 & 0.002* \\
 &  &  &  &  &  &  &  &  & CTX & S2S & 0.006 & 0.038* \\
 &  &  &  &  &  &  &  &  & HT & S2S & -0.002 & 0.567 \\
Word length frequencies & 0.066 & 0.001 & 0.066 & 0.001 & 0.066 & 0.001 & 0.497 & 0.611 & \multicolumn{4}{l}{--} \\
Common bigram frequencies & 0.026 & 0.000 & 0.026 & 0.001 & 0.026 & 0.001 & 0.048 & 0.953 & \multicolumn{4}{l}{--} \\
Common trigram frequencies & 0.050 & 0.001 & 0.050 & 0.001 & 0.050 & 0.001 & 0.138 & 0.871 & \multicolumn{4}{l}{--} \\
\hline
\end{tabular}
\begin{tablenotes}
\small
\item[*] Note: p<0.05
\end{tablenotes}
\end{threeparttable}
\end{sidewaystable}

\begin{table}
\centering
\begin{threeparttable}
\caption{Excerpt from Book 22}
\begin{tabular}{>{\raggedright\arraybackslash}p{0.3\linewidth}>{\raggedright\arraybackslash}p{0.6\linewidth}}
\hline
\textbf{Translations} & \textbf{Sample text (22.chs re.txt)} \\
\hline
ST & \begin{CJK*}{UTF8}{gbsn}从他的角度，透过圆弧形的金鱼屋檐，可以看到被挡住太阳了的蓝天。路胜学着他的样子，坐到一侧的另一张靠椅上。椅子是竹制的，躺上去很冰凉。微风习习，从花园里吹拂进来，带出丝丝花香。\end{CJK*} \\
HT & [From his angle, he could \textcolor{red}{peer through} the arched goldfish-patterned roof, which blocked the sun, towards \textcolor{red}{the azure blue sky}.] Lu Sheng \textcolor{red}{imitated his posture}, sitting down on another bench next to him. The bench \textcolor{red}{was woven of rattan} and was cold to the touch. \textcolor{blue}{A breeze blew into} the pavilion, carrying with it a \textcolor{blue}{floral fragrance}.  \\
S2S & [From his angle, through the arc of the goldfish house eaves, he could \textcolor{red}{see the blue sky} that blocked the sun.] Lu Sheng, \textcolor{red}{imitating him}, sat on another chair to the side. The chair \textcolor{red}{was made of bamboo}, and it was very cold to lie on. \textcolor{blue}{A gentle breeze blew in} from the garden, bringing with it \textcolor{blue}{the scent of flowers}. \\
CTX & [From his angle, through the arc-shaped eaves of the goldfish house, he could \textcolor{red}{see the blue sky} that blocked the sun.] Lu Sheng, \textcolor{red}{following his example}, sat on another recliner on the side. The chair \textcolor{red}{was made of bamboo}, and it was very cool to lie on. \textcolor{blue}{A gentle breeze blew in} from the garden, bringing with it a hint of \textcolor{blue}{floral fragrance}. \\
\hline

\end{tabular}
\begin{tablenotes}
\small
\item[*]Note: Similar vocabulary in blue; different vocabulary in red.
\end{tablenotes}
\end{threeparttable}
\end{table}

\begin{sidewaystable}
\caption{ANOVA results for syntactic features}
\begin{tabular}{ll l l l l l l l l l l l}
\hline
\multirow{2}{*}{\textbf{Syntactic features}} & \multicolumn{2}{c}{\textbf{HT}} & \multicolumn{2}{c}{\textbf{S2S}} & \multicolumn{2}{c}{\textbf{CTX}} & \multirow{2}{*}{\textbf{F}} & \multirow{2}{*}{\textbf{Sig.}} & \multicolumn{4}{l}{\textbf{Pair-wise comparison}} \\
 & M & Std. & M & Std. & M & Std. &  &  & Group 1 & Group 2 & Mean dif. & Sig. \\
\hline
Sentence length & 17.196 & 3.601 & 15.493 & 2.279 & 17.416 & 3.558 & 2.697 & 0.074 & \multicolumn{4}{l}{--} \\
Function words frequencies & 0.007 & 0.009 & 0.007 & 0.009 & 0.007 & 0.009 & 0.051 & 0.950 & \multicolumn{4}{l}{--} \\
Transition words frequencies & 0.000 & 0.001 & 0.000 & 0.000 & 0.000 & 0.000 & 0.704 & 0.495 & \multicolumn{4}{l}{--} \\
Punctuation character frequencies & 0.111 & 0.001 & 0.111 & 0.001 & 0.111 & 0.001 & 0.154 & 0.858 & \multicolumn{4}{l}{--} \\
\multirow{3}{*}{Simple sentence percentage} & \multirow{3}{*}{0.584} & \multirow{3}{*}{0.083} & \multirow{3}{*}{0.542} & \multirow{3}{*}{0.058} & \multirow{3}{*}{0.519} & \multirow{3}{*}{0.074} & \multirow{3}{*}{5.265} & \multirow{3}{*}{0.007*} & CTX & HT & 0.066 & 0.006* \\
 &  &  &  &  &  &  &  &  & CTX & S2S & 0.023 & 0.498 \\
 &  &  &  &  &  &  &  &  & HT & S2S & -0.042 & 0.104 \\
\multirow{3}{*}{Complex sentence percentage} & \multirow{3}{*}{0.416} & \multirow{3}{*}{0.083} & \multirow{3}{*}{0.458} & \multirow{3}{*}{0.058} & \multirow{3}{*}{0.481} & \multirow{3}{*}{0.074} & \multirow{3}{*}{5.265} & \multirow{3}{*}{0.007*} & CTX & HT & -0.066 & 0.006* \\
 &  &  &  &  &  &  &  &  & CTX & S2S & -0.023 & 0.498 \\
 &  &  &  &  &  &  &  &  & HT & S2S & 0.042 & 0.104 \\
\hline
\end{tabular}

\end{sidewaystable}

\begin{table}
\centering
\begin{threeparttable}
\caption{Excerpt from Book 296}
\begin{tabular}{>{\raggedright\arraybackslash}p{0.2\linewidth}>{\raggedright\arraybackslash}p{0.6\linewidth}}
\hline
\textbf{Translations} & \textbf{Sample text (296.chs\_re.txt)} \\
\hline
ST & … \begin{CJK*}{UTF8}{gbsn}那药香味若有若无，但随着他渐渐走近药香便越来越浓。… 正是五香芝的气味儿。他循着气味前行，来到一座山崖边，俯身向下看去，只见山崖陡峭，乱石突出，谷中幽深，深不见底。\end{CJK*} \\
HT & It was faint and vague, but as he walked closer, the fragrance of the spirit herb became thicker and stronger. … This was the unique smell of the Five-Fragrance Lingzhi Mushroom! He followed behind and tracked the unique smell to the side of a cliff and \textcolor{blue}{he slanted his body to look down. The cliff was steep and filled with protruding rocks. The valley was so deep that not even his eyes or the light could reach to the bottom}.  \\
S2S & \textcolor{red}{The scent was faint yet distinct, growing stronger as he approached its source}. … - the signature scent of the Five-Spice Ganoderma. \textcolor{red}{Following the scent, he arrived at the edge of a cliff}. \textcolor{blue}{Looking down, he saw a steep precipice, protruding rocks, and a deep, unfathomable valley}.  \\
CTX & \textcolor{red}{Zhong Yue traversed mountains and ridges, suddenly catching a faint whiff of a peculiar medicinal fragrance}. The fragrance was faint but grew stronger as he approached. …, which were the scents of the Wuxiang Zhi. He followed the scent and came to the edge of a cliff. \textcolor{blue}{Looking down, he saw the cliff was steep, with protruding rocks and a deep valley that was bottomless}.  \\
\hline

\end{tabular}
\begin{tablenotes}
\small
\item[*]Note: Texts highlighted in blue for comparison across three translations. Participle clauses in GPT translations are highlighted in red.
\end{tablenotes}
\end{threeparttable}
\end{table}
 
\begin{table}
\centering
\begin{threeparttable}
\caption{Excerpt from Book 168}
\begin{tabular}{>{\raggedright\arraybackslash}p{0.15\linewidth}>{\raggedright\arraybackslash}p{0.6\linewidth}}
\hline
\textbf{Translations} & \textbf{Sample text (168.chs\_re.txt)} \\
\hline
ST & \begin{CJK*}{UTF8}{gbsn}唯一剩下的一扇门上写着“灵霄殿”三个字，此时正处于虚掩的状态。纥溪缓缓推开房门，脑中瞬间涌入一个模糊的信息——“九重天宫，此为灵霄，万物生灵，皆为汝用。\end{CJK*}”  \\
HT & There was now only one \textcolor{red}{door} remaining with the \textcolor{red}{words “Spirit Firmament Door”} on top of it, and at this \textcolor{red}{momen}t it was closed. \textcolor{red}{Hexi} slowly pushed the \textcolor{red}{door} open, and in a \textcolor{red}{flash}, her \textcolor{red}{mind} was suddenly filled with vague \textcolor{red}{information}— “Ninth \textcolor{red}{Palace} in \textcolor{red}{Heaven}, this is \textcolor{red}{Spirit Firmament}. All living \textcolor{red}{things} are for thou to use.”  \\
S2S & The only remaining \textcolor{red}{door} had the \textcolor{red}{words 'Lingxiao Palace'} written on it, and it was slightly ajar. \textcolor{red}{Ge Xi} slowly pushed open the \textcolor{red}{door}, and a vague \textcolor{red}{piece} of \textcolor{red}{information} flooded into her \textcolor{red}{mind} - 'Nine \textcolor{red}{Heavens Palace}, this is \textcolor{red}{Lingxiao}, all \textcolor{red}{creatures}, are for your \textcolor{red}{use}.'  \\
CTX & The only remaining \textcolor{red}{door} had the \textcolor{red}{words 'Lingxiao Palace'} written on it and was slightly ajar. \textcolor{red}{Ge Xi} slowly pushed open the \textcolor{red}{door}, and a vague \textcolor{red}{piece} of \textcolor{red}{information} instantly flooded into her \textcolor{red}{mind} - 'Nine \textcolor{red}{Heavens Palace}, this is \textcolor{red}{Lingxiao}, all \textcolor{red}{creatures}, are for your \textcolor{red}{use}.'  \\
\hline

\end{tabular}
\begin{tablenotes}
\small
\item[*]Note: Nouns are highlighted in red.
\end{tablenotes}
\end{threeparttable}
\end{table}

\end{document}